\documentclass{article}
\usepackage{ijcai19}

\usepackage{times}
\usepackage{soul}
\usepackage{url}
\usepackage[hidelinks,pageanchor=true,colorlinks,linkcolor=blue,citecolor=blue,urlcolor=blue]{hyperref}
\usepackage[utf8]{inputenc}
\usepackage[small]{caption}
\usepackage{graphicx}
\usepackage{amsmath}
\usepackage{booktabs}
\usepackage[dvipsnames]{xcolor}

\usepackage{natbib}

\usepackage{fancyvrb} 
\usepackage{multirow}
\usepackage[usestackEOL]{stackengine} 
\usepackage{bm} 

\title{Tackling Domain-Specific Winograd Schemas with Knowledge-Based Reasoning and Machine Learning}

\author{
Suk Joon Hong$^{1,2}$\footnote{Contact Author}\And
Brandon Bennett$^1$\\
\affiliations
$^1$University of Leeds\\
$^2$InfoMining Co.\\
\emails
{\tt hsjplus@infomining.co.kr}, \ \ \ 
{\tt B.Bennett@leeds.ac.uk}
}

\begin{document}

\maketitle

\begin{abstract}
The \emph{Winograd Schema Challenge\/} (WSC) is a common sense reasoning task that requires background knowledge. In this paper, we contribute to tackling WSC in four ways. Firstly, we suggest a keyword method to define a restricted domain where distinctive high-level semantic patterns can be found. A \emph{thanking domain} was defined by keywords, and the data set in this domain is used in our experiments. Secondly, we develop a high-level knowledge-based reasoning method using semantic roles which is based on the method of \cite{sharma:2019}. Thirdly, we propose an ensemble method to combine knowledge-based reasoning and machine learning which shows the best performance in our experiments. As a machine learning method, we used Bidirectional Encoder Representations from Transformers (BERT) \citep{kocijan:2019}. Lastly, in terms of evaluation, we suggest a `robust' accuracy measurement by modifying that of \cite{trichelair:2018}. As with their switching method, we evaluate a model by considering its performance on trivial variants of each sentence in the test set.
\end{abstract}

\section{Introduction}
The \emph{Winograd Schema Challenge\/} (WSC) was proposed by \cite{levesque:2012} as a means to test whether a 
machine has human-like intelligence. It is an alternative to the well known
\emph{Turing Test\/} (TT) and has been designed with the motivation of reducing certain problematic aspects
that affect the TT. Specifically, while the TT is subjective in nature, the WSC provides a purely
objective evaluation; and whereas
passing the TT requires a machine to behave in a deceptive way, the WSC takes the form of a positive demonstration
of intelligent capability.

The core problem of the WSC is to resolve the reference of pronouns occurring in natural language sentences. 
To reduce the possibility that the task can be accomplished by procedures based on superficial or
statistical characteristics, rather than `understanding' of the sentence,
they specify that the test sentences used in the WSC, should be constructed in pairs, which have similar
structure and differ only in some key word or phrase, and such that the correct referent of the pronoun
is different in the two cases. This sentence pair, together with an indication of which pronoun is to
be resolved and a pair of two possible candidates, is called a \emph{Winograd Schema}.

The following is an example of the Winograd schemas from the original WSC273 data set \citep{levesque:2012}: 

\begin{enumerate}
\item The trophy doesn't fit in the brown suitcase because {\bf it} is too {\em large\/}.
\begin{itemize}
\item {\bf Candidates for the pronoun:} the trophy / the suitcase, {\bf Answer:} the trophy
\end{itemize}
\item The trophy doesn't fit in the brown suitcase because {\bf it} is too {\em small\/}.
\begin{itemize}
\item {\bf Candidates for the pronoun:} the trophy / the suitcase, {\bf Answer:} the suitcase 
\end{itemize}
\end{enumerate}

\cite{levesque:2012} design Winograd schemas to require background knowledge to resolve a pronoun, which can be an evidence of {\em thinking\/}. Therefore, they exclude the sentences that can be resolved by a statistical association within a sentence. 

In this paper, we introduce a keyword method to define domains in Winograd schemas. To our best knowledge, this is the first work to use keywords for defining domains in WSC and explore high-level patterns in them. To use the domain-specific high-level patterns, we also develop an advanced high-level knowledge-based reasoning method by modifying the method of \cite{sharma:2019}. Furthermore, we suggest a simple ensemble method that combines knowledge-based reasoning and machine learning. By the experiments on the domain-specific data set, the ensemble method gives a better performance than each single method. Lastly, we also propose a `robust' accuracy 
measure that is more objective by improving the switching method of \cite{trichelair:2018}. 

\section{Related work}
Knowledge-based reasoning and machine learning are the two main approaches to resolve Winograd schemas.

\begin{table}
	\centering
	\begin{tabular}{c|p{3.9cm}|p{1.1cm}|p{1.1cm}}
	    \toprule
		Type & Sentence & Pred. & Answer\\ 
        \midrule
		Ori. & Dan had to stop Bill from toying with the injured bird. {\bf He} is very compassionate. &  {\bf Dan} & Dan \\
		Neg. & Dan had to stop Bill from toying with the injured bird. {\bf He} is {\bf not} compassionate. & {\bf Dan} & Bill \\
		\midrule
		Ori. & I can't cut that tree down with that axe; {\bf it} is too small. &  {\bf The tree} & The axe \\
		Neg. & I can't cut that tree down with that axe; {\bf it} is {\bf not} small. &  {\bf The tree} & The tree \\
		\bottomrule
	\end{tabular}
	\caption{Two Examples from WSC273 with each variant by negation }
	\label{tab:bert_negation}
\end{table} 

\paragraph{Machine learning} 
An early work by \cite{rahman:2012} extracts features of a WSC-like sentence by using background knowledge such as Google search counts and a large corpus, and these features are used to train the SVM ranker that gives the higher rank to the correct candidate. 

More recent ML approaches mostly  use a neural language model. \cite{trinh:2018} introduce an approach to use a neural language model to tackle Winograd schemas. After this, Bidirectional Encoder Representations from Transformers (BERT) \citep{devlin:2018}, which is a state-of-the-art language model, is also used for WSC. \cite{kocijan:2019} demonstrate that the BERT fine-tuned with the data set similar to Winograd schemas gives a better performance than the BERT without fine-tuning. In addition, \cite{sakaguchi:2020} give the accuracy of around $90\%$ on the original WSC273 by fine-tuning a variant of BERT with the larger data set (WinoGrande) which is also similar to Winograd schemas.

Despite the the high accuracy of BERT and other neural language model methods, some limitations have been found. Though many of the original Winograd schemas can be resolved by the language models, \cite{trichelair:2018} demonstrate that they often predict wrongly on simple variants of the original sentences. Specifically, when we switch the positions of the candidates, in most cases this means that the answer should also be switched. However, the language model methods frequently give the same prediction for the switched sentence as in the original sentence. We return to this matter of switching in Section 6. Their finding implies that the real understanding of the model cannot be guaranteed by accuracy only. Furthermore, \cite{ettinger:2020} also shows that the BERT does not seem to understand negation since BERT's predictions on the masked tokens of the negated sentences are likely to be similar to its predictions on the masked tokens of the non-negated sentences. 

The finding of \cite{ettinger:2020} is also supported by the experiments on some Winograd schema sentences from WSC273 that are negated by us in Table \ref{tab:bert_negation}. Though the answers should be changed on the negated Winograd schema sentences in this example, the BERT's predictions on them are still same as its predictions on the non-negated sentences.  
\paragraph{Knowledge-based reasoning.} It is advantageous that knowledge-based reasoning methods can give logical explanations for the answers of Winograd schemas by reasoning. \citet[p.18]{bailey:2015} use the ``correlation calculus'' for the reasoning by using different levels of background knowledge principles. In addition, \cite{sharma:2019} automates graphical representations of a sentence and the reasoning by using K-Parser \citep{sharma:2015a} and Answer Set Programming (ASP) \citep{gelfond:1988}.

However, knowledge-based reasoning methods also have limitations on automation and building a knowledge base that covers this \emph{general domain\/}. \cite{bailey:2015} do not give an automatic method to transform a natural language sentence into the form of first-order logic that they use. Though \cite{sharma:2015b} have an automation method to extract background knowledge, their method is based on using a search engine, which cannot guarantee acquiring necessary knowledge.

\section{Semantic Domains and Keywords}
Assuming that keywords are related to the high-level semantic meaning of a sentence, we introduce a keyword method in terms of identifying a domain in Winograd Schemas. To our best knowledge, our method is the first work to use keywords regarding a domain in Winograd schemas and examine high-level patterns in a domain. 

For the pilot study, we chose a {\em thanking\/} domain since the thanking domain has a distinctive semantics. The thanking domain contains the sentences that have a keyword related to the normal sense of thanking. The keywords we used for the thanking domain were ``thank'' and ``grateful''. We extracted sentences that include the two keywords from WinoGrande \cite{sakaguchi:2020} which has approximately $44$K Winograd schema sentences since WSC273 contains only $273$ sentences. In this extraction, we exclude the sentences including ``thanks to'' and ``thanks in no small part to'' though ``thank'' is within them. The reason for their exclusion is that their semantic meaning is related to causal relations, not thanking. 

As a result, the number of the extracted Winograd schema sentences was $171$ ($\approx0.39\%$ of the $44,000$ Winogrande sentences). We believe that the number of them is adequate as it is compatible with the number of the original WSC273's sentences which is $273$. These extracted sentences are considered to belong to the thanking domain, and we investigated the high-level reasoning patterns in the thanking domain. As shown in Table \ref{tab:five_patterns}, the five major high-level domain-specific reasoning patterns were found. As these patterns are from the thanking domain, they are related to the relationships of ``owing'' and ``being owed''. It is common that a person who is owing would thank or do good to someone who is owed. It is interesting that around $77\%$ ($132/171$) of the sentences in the thanking domain follow the only five major high-level patterns. Some of the other minor high-level patterns were also found in the thanking domain.

\begin{table}
\centering
\begin{tabular}{lp{6.5cm}}  
\toprule
Type & Sentence \\
\midrule
Pattern 1 & Candidate1 owes candidate2, and (so) pronoun is doing good \\
Pattern 2 & Candidate1 owes candidate2, and (so) pronoun is receiving good \\
Pattern 3 & Candidate1 does good to candidate2 because pronoun is owing \\
Pattern 4 & Candidate1 gives thanks to candidate2 because pronoun is being owed \\
Pattern 5 & Candidate1 gives thanks to candidate2 because pronoun is owing \\
\bottomrule
\end{tabular}
\caption{The five major high-level domain-specific reasoning patterns found in the thanking domain}
\label{tab:five_patterns}
\end{table}

In addition to the high-level patterns, the Winograd schema sentences in the thanking domain have two other characteristics. The first characteristic is that more than $90\%$ ($161/171$) of the sentences in the thanking domain have candidates with human names while this proportion is around $50\%$ in WSC273. This finding can be explained by the fact that thanking is done by humans. For the second characteristic, only around $46\%$ ($80/171$) of the sentences in the thanking domain can be paired while almost all the sentences can be paired in WSC273. This is due to the fact that some of the WinoGrande sentences use keywords such as ``thank'' for the special words or the alternative words.  

\section{The advanced high-level knowledge-based reasoning method}
Our high-level knowledge-based reasoning method is related to the method of \cite{sharma:2019}, who
identifies and exploits very specific identity implications to
resolve pronouns. We use a more general method of abstracting semantic relationships to identify and make use of  high-level domain-specific semantic roles, based on the analysis of Winograd schemas given by \cite{bennett:2020}. According to this analysis, most Winograd sentences can be
represented as having the form:
\begin{equation}
\label{eqn:bennett}
\phi(a, b, p) \equiv ((\alpha(a)\ \land\ \beta(b)\ \land \rho(a, b))\ \#\ \pi(p))
\end{equation}
where $\alpha$ is the candidate $a$'s property, $\beta$ is the candidate $b$'s property, $\rho$ refers to a predicate that defines the candidates' relationship, $\#$ refers to the relationship between the part of the candidates and the part of the pronoun (e.g. ``because''), and $\pi$ is the pronoun $p$'s property. 

While this type of formula can be used for particular examples of Winograd schemas, we used the formula to represent higher-level general principles that can potentially explain a large class of specific cases. 

\subsection{Building a domain-specific knowledge base}
Our knowledge base is composed of two types of rules and one type of facts --- rules to derive semantic roles, rules to define relationships regarding the semantic roles and high-level background knowledge principles.

\paragraph{Rules to derive semantic roles} We defined rules to derive semantic roles specific to the thanking domain. These semantic roles are high-level representations related to the candidates and the pronoun, and they are also grounds to derive the relationships regarding them. In the thanking domain, six major domain-specific semantic roles were found --- thanker, being thanked, giver, given, helper and being helped. 
    
These rules are implemented in ASP by using K-Parser's graphical representations, and they are manually defined from the sentences in the thanking domain. For example, a simple rule for thanker can be defined as: 
\begin{Verbatim}
has_s(X, semantic_role, thanker) :- 
	has_s(Thank,agent,X), 
	has_s(Thank,instance_of,thank).
\end{Verbatim}
    
In order to make more generalisable rules, the following four measures were taken. The first measure is to derive the semantic role of a candidate if that of the other candidate is known (e.g. if ``give'' is the semantic role of a candidate, then that of the other candidate would be ``given''). The second measure is for the case when no semantic roles of the candidates are known. For instance, if candidate1 is an agent of the verb to which candidate2 is a recipient, candidate1's semantic role is derived to be ``giver''. The third measure is to use synonyms that are manually defined in the thanking domain. The fourth measure is to use an external sentiment lexicon dictionary \citep{hu:2004} to derive the semantic roles of ``good'' and ``bad''.

\begin{table}
	\centering
	\begin{tabular}{cccc}
	\toprule
		\rule{0pt}{3ex} \multirow{2}{1cm}{Semantic relationship} & \multirow{2}{1cm}{Causal relation} & \multicolumn{2}{c}{Semantic role}  \\\cline{3-4}
		\rule{0pt}{3ex}				      				  &    & X  & Y \\
	\midrule
		\rule{0pt}{3ex} X owes Y & No & being helped & helper\\
		\rule{0pt}{3ex} X owes Y & No & given & giver\\
		\rule{0pt}{3ex} X does good to Y & Yes & helper & being helped\\ 
		\rule{0pt}{3ex} X does good to Y & Yes & giver & given\\
		\rule{0pt}{3ex} X gives thanks to Y & Yes & thanker & being thanked\\
	\bottomrule
	\end{tabular}
	\caption{The major rules to define the relationships between the semantic roles of the candidates}
	\label{tab:semantic_relationship}
\end{table} 

\paragraph{Rules to define relationships regarding the semantic roles} The domain-specific semantic roles are used to derive their relationships for the high-level representations of Winograd schema sentences. We defined the rules for the relationships using the semantic roles in the following three aspects: relationships between the semantic roles of the candidates, relations between the candidates' part and the pronoun part, and property of the pronoun. 

\begin{enumerate}
    \item {\bf Relationships between the candidates' semantic roles}: As the five high-level patterns in Table \ref{tab:five_patterns} show, the two candidates in a Winograd schema are found to have domain-specific relationships in the thanking domain. The main relationships between them are ``owes'', ``does good to'' and ``gives thanks to''. In order to derive the relationships between the semantic roles of the candidates, we defined the rules by using their semantic roles and the existence of causal relation. Table \ref{tab:semantic_relationship} shows the five rules to derive the relationships between the candidates. 

    \hspace{5mm} For instance, the second rule in Table \ref{tab:semantic_relationship} means that if the semantic role of $X$ is ``given'', that of $Y$ is ``giver'', and there is no causal relation then $X$ owes $Y$. It is written in ASP as: 
    \begin{Verbatim}
has_s(X, owes, Y) :-
    has_s(X,semantic_role,given), 
    has_s(Y,semantic_role,giver), 
    not has_s(_,relation,causer).
    \end{Verbatim}
    
    \item {\bf A Relationship between the candidates' part and the pronoun part} As represented in the formula \ref{eqn:bennett}, we defined the rules to derive a relationship between the part of the candidates and the part of the pronoun (``$\#$''). In the thanking domain, a causal relationship is the only relationship we considered as the existence of the causal relationship affects the relationship between the candidates. If there is a causal relationship, the pronoun's part would explain the candidates' part, and if no causal relationship is derived, the candidates' part would explain the pronoun's part. These rules to derive the causal relationship were defined in ASP, and one of the rules is defined as:
    \begin{Verbatim}
has_s(P, relation, causer) :-
    pronoun(P),
    is_candidate(A),
    has_s(Verb1,caused_by,Verb2),
    1 {has_s(Verb1,agent,A);
    has_s(Verb1,recipient,A)},
    has_s(Verb2,agent,P).
    \end{Verbatim}    
    \item {\bf Property of the pronoun} The semantic role of the pronoun can be the property of the pronoun (``$\pi(p)$'') in the formula \ref{eqn:bennett}, but there can be a higher-level semantic role. For this reason, we defined the rules to derive the high-level semantic role from the low-level semantic role. These rules are based on the fact that a low-level semantic role can be {\em a subset of\/} a high-level semantic role in the thanking domain. For instance, the semantic role of ``helper" can be a subset of that of ``doing good''. We implemented these rules in ASP, and the following rule is one of them:
    \begin{verbatim}
has_s(X, semantic_role, doing_good) :-
    has_s(X, semantic_role, helper).
    \end{verbatim}
\end{enumerate}

\paragraph{High-level background knowledge principles} 
In our knowledge base, we also defined high-level domain-specific background knowledge principles as well as the two types of the rules above. The high-level background knowledge principles are used for the reasoning in comparison with the high-level representation of a sentence that is derived by the rules in the knowledge base. We followed the style of \cite{sharma:2019}'s background knowledge principles. 

\subsection{Transforming a Winograd schema sentence into a high-level representation}
We used K-Parser to transform the Winograd schema sentences in the thanking domain into the graphical representations as \cite{sharma:2019} does. By using the rules to derive semantic roles and to derive relationships between the semantic roles, we transformed the graphical representations into high-level representations. The following is an example of the transformations from WinoGrande:

\begin{center}
{\bf Kayla} cooked sticky white rice for {\bf Jennifer}, and [{\bf she}] was thanked 
for making such delicate rice.
\begin{itemize}
\item {\bf The semantic roles:} 
\begin{enumerate}
    \item Kayla: giver
    \item Jennifer: given 
    \item she: being thanked
\end{enumerate}
\item {\bf The relationships regarding the semantic roles:} 
\begin{enumerate}
    \item Jennifer {\bf owes} Kayla
    \item no causal relation 
    \item she is receiving good
\end{enumerate}
\end{itemize}
\end{center}

\subsection{Reasoning to derive the answer}
We used the reasoning rules of \cite{sharma:2019} with small modifications to resolve the Winograd schemas in the thanking domain. The goals of the modifications were to use the K-Parser outputs and the semantic roles for the reasoning. 

In the reasoning process, each Winograd schema sentence is compared with each background knowledge principle. As a result, the answer for each sentence can be a single answer, ``no answer'' and multiple answers. If multiple answers have the same answers, this case is considered as a single answer.   

\begin{figure*}
    \centering
	\centerline{\includegraphics{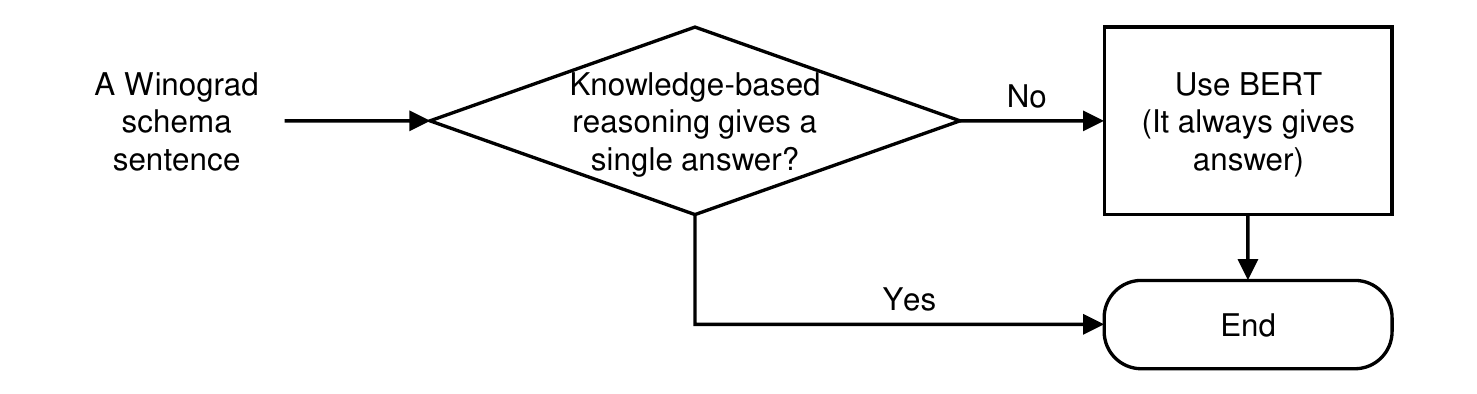}}
	\caption{Our algorithmic flow of combining the knowledge-based reasoning method and the machine learning method}
	\label{fig:combining_flow}
\end{figure*}

\section{The simple ensemble method}
We combined our advanced high-level knowledge-based reasoning method with BERT \citep{kocijan:2019}. The aim of our ensemble method is to mitigate each method's weakness. The weakness of the advanced high-level knowledge-based reasoning method is that if there are no rules that can be applied in the knowledge base, no answer can be derived. With respect to weakness of the BERT, its predictions are vulnerable to the small changes since it is not based on a logical relationship \citep{trichelair:2018,ettinger:2020}.

As shown in Figure \ref{fig:combining_flow}, we implemented a simple but effective ensemble method. If the knowledge-based reasoning method gives a single answer, the final answer would be this answer. On the other hand, if the prediction of the knowledge-based reasoning method is multiple answers or no answer, we would use the BERT's prediction for the final answer. With these two conditions, the weakness of each method can be reduced.    

\section{`Robust' accuracy}
As mentioned in Section~2, machine learning methods can give the incorrect answer on trivial
variants of sentences obtained by switching the candidates \citep{trichelair:2018}. 
This reveals an apparent weakness in these methods and a limitation in the simple evaluation of accuracy. 
Accuracy measurement is already quite tolerant because, since the number of the candidates are only two, the chance of predicting correctly without understanding is $50\%$. This is a further motivation for having a stricter form of accuracy measurement.
We propose a `robust' accuracy measurement based on a generalisation of \cite{trichelair:2018}. In addition to the switching, we add three more variants of each sentence by replacing the name of each candidate with the random name with the same gender if the candidates are both names. This basic method of replacing names should not affect the fundamental meaning of a sentence, and thus a model's incorrect predictions on the sentences where only the names are replaced can reveal its obvious lack of understanding. The following is an original sentence from WinoGrande in the thanking domain and its variants to measure the robust accuracy: 

\begin{itemize}
\item {\bf Original sentence}: {\bf Kayla} cooked sticky white rice for {\bf Jennifer}, and [she] was thanked for making such delicate rice.

\item {\bf The nouns switched}: {\bf Jennifer} cooked sticky white rice for {\bf Kayla}, and [she] was thanked for making such delicate rice.

\item {\bf The nouns replaced 1}: {\bf Tanya} cooked sticky white rice for {\bf Kayla}, and [she] was thanked for making such delicate rice.

\item {\bf The nouns replaced 2}: {\bf Erin} cooked sticky white rice for {\bf Tanya}, and [she] was thanked for making such delicate rice.

\item {\bf The nouns replaced 3}: {\bf Lindsey} cooked sticky white rice for {\bf Christine}, and [she] was thanked for making such delicate rice.
\end{itemize}

Only when a model predicts correctly on all of the original Winograd schema sentence and the four variants including the switched one, that prediction is considered to be 'robustly' accurate.

While the probability of predicting correctly on both switched and non-switched sentences out of luck is $0.5 \times 0.5 = 0.25$, the probability can go down to $(0.5)^5 \approx 0.03$ in the robust accuracy. In this sense, our robust accuracy is more objective on evaluating a model's performance. The limitation of the robust accuracy is that the candidates should be human names to make variants. In the case of no human names for the candidates, we only used the switching method to make a variant. This kind of exception is not common in the thanking domain since more than $90\%$ of the sentences have the candidates with human names. 

\section{Evaluation}
Our evaluation compares the performance of the following methods: our advanced high-level knowledge-based reasoning method, BERT \citep{kocijan:2019}, the BERT fine-tuned with the train set in the thanking domain and our ensemble method. The version of BERT \citep{kocijan:2019} we used is their best performing model (``BERT\_WIKI\_WSCR'') implemented in PyTorch in their repository, so the results of it can be replicated. They were evaluated on the $80$ paired Winograd schema sentences in the thanking domain, and the $91$ non-paired sentences were used for validation. For the evaluation metrics, we used accuracy and the robust accuracy.   

We did two experiments with the paired sentences in the thanking domain. In the first experiment, each pair was {\em separated \/} into either train set or test set. By its definition, $50\%$ of the paired sentences were used for the train set, and the others were used for the test set. In the second experiment, on the other hand, each pair was put {\em together} into one of them in a random manner. Considering the small number of the data set and the balance with the first experiment, the second experiment also took the $50:50$ split between the train set and the test set. 

\subsection{Results}
Tables \ref{tab:experiment1} and \ref{tab:experiment2} show the results of the two experiments respectively. Some same patterns were found in both experiments. The accuracies and the robust accuracies of our ensemble model are better than those of the other methods. Also, the models that contain the BERT were found to have the lower robust accuracies than the accuracies. It demonstrates that the BERT, as a machine learning method, can be weak to minor changes.

Different patterns were also found between the two experiments. The accuracy of the knowledge-based reasoning method in the first experiment was higher than that in the second experiment by a large margin. It implies that the close similarity between the train set and the test set is advantageous for the knowledge-based reasoning method since the rules defined by the train set are expected to be used for the test set. 

On the other hand, the fine-tuned BERT gave the opposite results since the better accuracy was found in the second experiment, not in the first experiment. This result can be explained by the characteristics of Winograd schemas. While similar sentences have different answers in a Winograd schema, the BERT is likely to give the same answer with that of the similar sentence, which leads to the wrong predictions in the first experiment. This result is compatible with the finding of \cite{kocijan:2019} that training with the paired sentences shows a better performance than training with the non-paired sentences.

\begin{table}
	\centering
	\begin{tabular}{p{12em}|p{4em}|p{4em}}
	    \toprule
		\centerline{Model} & Accuracy & \Longstack{`Robust'\\accuracy} \\ 
		\midrule
                \\[-0.7ex]
		\centerline{BERT} \centerline{\cite{kocijan:2019}} &
                \centerline{$70.0\%$} \centerline{$(28/40)$}  &
                \centerline{$62.5\%$} \centerline{$(25/40)$} \\
		\centerline{BERT fine-tuned}  \centerline{with the train set} & 
                \centerline{$47.5\%$} \centerline{$(19/40)$} &
                \centerline{$42.5\%$} \centerline{$(17/40)$} \\
		\centerline{Our knowledge-based} \centerline{reasoning
                  method} & 
                \centerline{$72.5\%$} \centerline{$(29/40)$} &
                \centerline{$72.5\%$} \centerline{$(29/40)$} \\
		\centerline{Our knowledge-based} \centerline{reasoning
                  method +}
                \centerline{BERT \cite{kocijan:2019}} & 
                \centerline{\boldsymbol{$90.0\%$}}
                \centerline{$(36/40)$} & 
                \centerline{\boldsymbol{$85.0\%$}} \centerline{$(34/40)$} \\ 
		\bottomrule
	\end{tabular}
	\caption{The results of the first experiment}
	\label{tab:experiment1}
\end{table}

\begin{table}
	\centering
	\begin{tabular}{p{12em}|p{4em}|p{4em}}
	    \toprule
		\centerline{Model} & Accuracy & \Longstack{`Robust'\\accuracy} \\ 
		\midrule
                \\[-0.7ex]
		\centerline{BERT} \centerline{\cite{kocijan:2019}} &
                \centerline{$77.5\%$} \centerline{$(31/40)$}  &
                \centerline{$70\%$} \centerline{$(28/40)$} \\
		\centerline{BERT fine-tuned}  \centerline{with the train set} & 
                \centerline{$75.0\%$} \centerline{$(30/40)$} &
                \centerline{$70.0\%$} \centerline{$(28/40)$} \\
		\centerline{Our knowledge-based} \centerline{reasoning
                  method} & 
                \centerline{$37.5\%$} \centerline{$(15/40)$} &
                \centerline{$37.5\%$} \centerline{$(15/40)$} \\
		\centerline{Our knowledge-based} \centerline{reasoning
                  method +}
                \centerline{BERT \cite{kocijan:2019}} & 
                \centerline{\boldsymbol{$80.0\%$}}
                \centerline{$(32/40)$} & 
                \centerline{\boldsymbol{$72.5\%$}} \centerline{$(29/40)$} \\ 
		\bottomrule
	\end{tabular}
	\caption{The results of the second experiment}
	\label{tab:experiment2}
\end{table}


\section{Conclusion}
This paper demonstrates that combining both the high-level knowledge-based reasoning method and the BERT can give a better performance in the thanking domain.

In this paper, we also introduced the keywords method to identify a domain, and this method can be applied to specify other domains. We showed that high-level patterns were found in the domain defined by the keywords. As only one domain --- the thanking domain --- was tackled, future work needs to be done with more domains in Winograd schemas. Though the number of the thanking domain is $171$ (around $0.39\%$ of the number of the WinoGrande) as a pilot study, some other domains could be larger than the thanking domain. For instance, the domain that can be defined by the keywords ``love'' and ``hate'' has $1,351$ (around $3\%$) and $612$ (around $1\%$) sentences respectively. If these were genuinely separate domains and the correct resolution of
each schema were based on principles in the domain corresponding to the key words it contains, this would imply that tackling around $100$ domains could cover almost all domains in Winograd schemas.

By modifying the method of \cite{sharma:2019} and focusing on the domain-specific semantic roles, we were able to develop a knowledge-based reasoning method that can use domain-specific high-level patterns. Though our knowledge-based method uses background knowledge principles that are built manually, we believe that our principles are more accurate than the kinds of semantic feature that could be reliably extracted from a large corpus or by using a search engine. This is because the simple statistical
method used for automatically extracting knowledge is vulnerable to data bias or special usage of words in idioms (e.g. ``thanks to'' referring to causal relations that do not involve thanking in the normal sense of this concept). In addition, our knowledge-based method can also be used in other natural language tasks such as Choice Of Plausible Alternaties (COPA) \citep{roemmele:2011}. But K-Parser used in our approach still needs to be improved as manual corrections were needed in some cases.

We also propose the robust accuracy by improving the method of \cite{trichelair:2018}. The decreased robust accuracies of the BERT reveal that its accuracy may not entail its real understanding.  

\subsection*{Code repository}
The code for the advanced high-level knowledge-based reasoning method (described in Section 4) can be accessed from the following repository: {\tt\url{https://github.com/hsjplus/high-level-kb-reasoning}}

\bibliographystyle{named}
\bibliography{main}

\begin{thebibliography}{}

\bibitem[\protect\citeauthoryear{Bailey \bgroup \em et al.\egroup
  }{2015}]{bailey:2015}
Daniel Bailey, Amelia Harrison, Yuliya Lierler, Vladimir Lifschitz, and Julian
  Michael.
\newblock The winograd schema challenge and reasoning about correlation.
\newblock In {\em 2015 AAAI Spring Symposium Series}, USA, 2015.

\bibitem[\protect\citeauthoryear{Bennett}{2020}]{bennett:2020}
Brandon Bennett.
\newblock Logical analysis of winograd schemas.
\newblock {\em Unpublished}, 2020.

\bibitem[\protect\citeauthoryear{Devlin \bgroup \em et al.\egroup
  }{2018}]{devlin:2018}
Jacob Devlin, Ming~W. Chang, Kenton Lee, and Kristina Toutanova.
\newblock Bert: Pre-training of deep bidirectional transformers for language
  understanding.
\newblock {\em arXiv:1810.04805[cs.CL]}, 2018.

\bibitem[\protect\citeauthoryear{Ettinger}{2020}]{ettinger:2020}
Allyson Ettinger.
\newblock What bert is not: Lessons from a new suite of psycholinguistic
  diagnostics for language models.
\newblock {\em Transactions of the Association for Computational Linguistics},
  8:34--48, 2020.

\bibitem[\protect\citeauthoryear{Gelfond and Lifschitz}{1988}]{gelfond:1988}
Michael Gelfond and Vladimir Lifschitz.
\newblock The stable model semantics for logic programming.
\newblock In {\em Proceedings of International Logic Programming Conference and
  Symposium}, pages 1070--1080, 1988.

\bibitem[\protect\citeauthoryear{Hu and Liu}{2004}]{hu:2004}
Minqing Hu and Bing Liu.
\newblock Mining and summarizing customer reviews.
\newblock In {\em Proceedings of the ACM SIGKDD International Conference on
  Knowledge Discovery and Data Mining (KDD2004)}, 2004.

\bibitem[\protect\citeauthoryear{Kocijan \bgroup \em et al.\egroup
  }{2019}]{kocijan:2019}
Vid Kocijan, Ana~M. Cretu, Oana~M. Camburu, Yordan Yordanov, and Thomas
  Lukasiewicz.
\newblock A surprisingly robust trick for winograd schema challenge.
\newblock In {\em Proceedings of the 57th Annual Meeting of the Association for
  Computational Linguistics}, pages 4837--–4842, 2019.

\bibitem[\protect\citeauthoryear{Levesque \bgroup \em et al.\egroup
  }{2012}]{levesque:2012}
Hector Levesque, Ernest Davis, and Leora Morgenstern.
\newblock The winograd schema challenge.
\newblock In {\em The 13th International Conference on Principles of Knowledge
  Representation and Reasoning}, Italy, June 2012.

\bibitem[\protect\citeauthoryear{Rahman and Ng}{2012}]{rahman:2012}
Altaf Rahman and Vincent Ng.
\newblock Resolving complex cases of definite pronouns: The winograd schema
  challenge.
\newblock In {\em EMNLP-CoNLL}, 2012.

\bibitem[\protect\citeauthoryear{Roemmele \bgroup \em et al.\egroup
  }{2011}]{roemmele:2011}
Melissa Roemmele, Cosmin~A. Bejan, and Andrew~S. Gordon.
\newblock Choice of plausible alternatives: An evaluation of commonsense causal
  reasoning.
\newblock In {\em AAAI Spring Symposium on Logical Formalizations of
  Commonsense Reasoning}, USA, March 2011.

\bibitem[\protect\citeauthoryear{Sakaguchi \bgroup \em et al.\egroup
  }{2020}]{sakaguchi:2020}
Keisuke Sakaguchi, Ronan~Le Bras, Chandra Bhagavatula, and Yejin Choi.
\newblock Winogrande: An adversarial winograd schema challenge at scale.
\newblock In {\em AAAI-20}, 2020.

\bibitem[\protect\citeauthoryear{Sharma \bgroup \em et al.\egroup
  }{2015a}]{sharma:2015a}
Arpit Sharma, Nguyen~H. Vo, Somak Aditya, and Chitta Baral.
\newblock Identifying various kinds of event mentions in k-parser output.
\newblock In {\em Proceedings of the The 3rd Workshop on EVENTS: Definition,
  Detection, Coreference, and Representation}, pages 82--88. Association for
  Computational Linguistics, 2015.

\bibitem[\protect\citeauthoryear{Sharma \bgroup \em et al.\egroup
  }{2015b}]{sharma:2015b}
Arpit Sharma, Nguyen~H. Vo, Somak Aditya, and Chitta Baral.
\newblock Towards addressing the winograd schema challenge - building and using
  a semantic parser and a knowledge hunting module.
\newblock In {\em IJCAI 2015}, pages 1319--1325, 2015.

\bibitem[\protect\citeauthoryear{Sharma}{2019}]{sharma:2019}
Arpit Sharma.
\newblock Using answer set programming for commonsense reasoning in the
  winograd schema challenge.
\newblock {\em arXiv:1907.11112[cs.AI]}, 2019.

\bibitem[\protect\citeauthoryear{Trichelair \bgroup \em et al.\egroup
  }{2018}]{trichelair:2018}
Paul Trichelair, Ali Emami, Adam Trischler, Kaheer Suleman, and Jackie~C.~K.
  Cheung.
\newblock How reasonable are common-sense reasoning tasks: A case-study on the
  winograd schema challenge and swag.
\newblock {\em arXiv:1811.01778[cs.LG]}, 2018.

\bibitem[\protect\citeauthoryear{Trinh and Le}{2018}]{trinh:2018}
Trieu~H. Trinh and Quoc~V. Le.
\newblock A simple method for commonsense reasoning.
\newblock {\em arXiv:1806.02847[cs.AI]}, 2018.

\end{thebibliography}

\end{document}